# Performance Comparison of SVM and ANN for Handwritten Devnagari Character Recognition

Sandhya Arora[1]. Debotosh Bhattacharjee[2], Mita Nasipuri[2], L. Malik[4], M. Kundu[2] and D. K. Basu[3]

[1] Dept. of CSE & IT, Meghnad Saha Institute of Technology
kolkata, 700150,India
[1] *sandhyabhagat@yahoo.com*

[2] Department of Computer Science and Engineering, Jadavpur University
Kolkata, 700032,India

[3] AICTE Emeritus Fellow, Department of Computer Science and Engineering, Jadavpur University
Kolkata, 700032,India

[4] Dept. of Computer science. G.H. Raisoni college of Engineering
Nagpur, India

**Abstract**
Classification methods based on learning from examples have been widely applied to character recognition from the 1990s and have brought forth significant improvements of recognition accuracies. This class of methods includes statistical methods, artificial neural networks, support vector machines (SVM), multiple classifier combination, etc. In this paper, we discuss the characteristics of the some classification methods that have been successfully applied to handwritten Devnagari character recognition and results of SVM and ANNs classification method, applied on Handwritten Devnagari characters. After preprocessing the character image, we extracted shadow features, chain code histogram features, view based features and longest run features. These features are then fed to Neural classifier and in support vector machine for classification. In neural classifier, we explored three ways of combining decisions of four MLP's, designed for four different features.
***Keywords***: *Support Vector Machine, Neural Networks, Feature Extraction, Classification.*

## 1. Introduction

Over the years, computerization has taken over large number of manual operations, one such example is off-line handwritten character recognition, which is the ability of a computer system to receive and interpret handwriting input present in the form of scanned images. In the early stage of OCR (optical character recognition) development, template matching based recognition techniques were used [16]. The templates or prototypes in these early methods were designed artificially, selected or averaged from few samples. As the number of samples increased, this simple design methodology, became insufficient to accommodate the shape variability of samples, and so, are not able to yield high recognition accuracies. To take full advantage of large volume of sample data, the character recognition community has turned attention to classification methods based on learning from examples strategy, especially based on artificial neural networks (ANNs) from the late 1980s and the 1990s. New learning methods, using support vector machines (SVMs), are now actively studied and applied in pattern recognition problems. Learning methods have beneficiated character recognition methods tremendously. They relieve us from painful job of template selection and tuning, and the recognition accuracies get improved significantly because of learning from large sample data. Some excellent results have been reported [17, 18, 19]. Despite the improvements, the problem is far from being solved. The recognition accuracies of either machine-printed characters on degraded document image or freely handwritten characters are still insufficient.

The recent spurt in the advancement in handwriting recognition has provided publications but do not involve the performance comparison of artificial neural networks and support vector machines on the same feature set for handwritten Devnagari characters. In this paper, we discuss the results of ANNs and SVM applied on handwritten Devnagari Characters. The strengths and weaknesses of these classification methods will also be discussed.

## 2. Challenges in Handwritten Devnagari Recognition

The Devanagari script has descended from the Brahmi script sometime around the 11th century AD. It was originally developed to write Sanskrit but was later adapted to write many other languages like Bhojpuri, Bhili, Magahi, Maithili, Marwari, Newari, Pahari, Santhali, Tharu, Marathi, Mundari, Nepali and Hindi.

Hindi, is the official national language of India and also the third most popular language in the world. According to a recent survey, Hindi is being used by 551 million people in India.

The basic set of symbols of Devnagari script consists of 36 consonants (or *vyanjan*) and 13 vowels (or *swar*) as shown in Figure 1. The characters may also have *half* forms. A half character in most of the cases is touched by the following character, resulting in a composite character , also known as compound character. The script has a set of modifier symbols which represent the modified shapes undertaken by the vowels, when they are combined with consonants, as own in Figure 2. These symbols are placed either on top, at the bottom, on the left, to the right or a combination of these. There are infinite variations of handwriting of individuals because of perceptual variability and generative variability. This variability effects in handwriting make the machine recognition of handwritten characters difficult. OCR of Devnagari script documents becomes further complicated due to the presence of compound characters and modifiers that make character separation and identification very difficult. All the individual characters are joined by a head line called "*Shiro Rekha*" in case of Devanagari Script. This makes it difficult to isolate individual characters from the words. There are various isolated dots, which are vowel modifiers, namely, "*Anuswar*", "*Visarga*" and "*Chandra Bindu*", which add up to the confusion. Ascenders and Descender recognition is also complex.

Figure 1: Vowels and Consonants of Devnagari Script

Figure 2: Vowels and corresponding modifiers of Devnagari Script

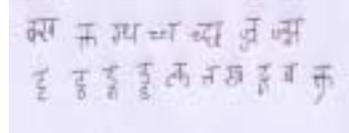

Figure 3: Combination of half consonant and consonant (compound characters)

## 3. State of the Art

Here, we will confine the discussion strictly on, feature based classification methods for off-line Devnagari character recognition. These methods can be grouped into four broad categories namely, statistical methods, ANN based methods, Kernel based methods, and multiple classifier combination, which are discussed below. We discuss the work done on Devnagari characters, using some of these classification methods.

### 3.1 Statistical methods

Statistical classifiers are rooted in the Bayes decision rule, and can be divided into parametric ones and non-parametric ones [20, 21]. Non-parametric methods, such as Parzen window, the nearest neighbor (1-NN) and k-NN rules, the decision-tree, the subspace method, etc, are not much used, since all training samples are stored and compared. Sinha and Bansal [1,2] have explored various knowledge sources at all levels. Initial segmentation process use horizontal and vertical histogram for line and word separation. Horizontal zero crossings, moments, aspect ratio, pixel density in 9 zones, number and position of vertex points and structural description of character are taken as classifying features. Dictionary is used at post processing step. Sethi and Chatterjee [7] proposed a decision tree based approach for recognition of constrained hand printed Devnagari characters using primitive features.

Parametric classifiers include the linear discriminant function (LDF), the quadratic discriminant function (QDF), the Gaussian mixture classifier, etc. An improvement to QDF, named regularized discriminant analysis (RDA), was shown to be effective to overcome inadequate sample size [50] and stabilizes the performance of QDF by smoothing the covariance matrices[25]. The modified quadratic discriminant function (MQDF) proposed by Kimura et al. was shown to improve the accuracy, memory, and computation efficiency of the QDF [22]. They used directional information obtained from arc tangent of the gradient. The directions are sampled down using Gaussian filter to get 392 dimensional feature-vector. This feature vector is applied on MQDF classifier. For modeling multi-modal distributions, the mixture of Gaussians in high dimensional feature space does not necessarily give high classification accuracy, yet the mixture of linear subspaces has shown effects in handwritten character recognition [23, 24]. R. Kapoor, D. Bagai, T. Kamal, [6] proposed HMM based approach, using junction points of a character as the main feature. The character has been divided into three major zones. Three major features i.e. number of paths, direction of paths, and region of the node were extracted from the middle zone.

### 3.2 Artificial neural networks

Feedforward neural networks, including multilayer perceptron (MLP) [51], radial basis function (RBF) network [52], the probabilistic neural network (PNN) [53], higher-order neural network (HONN) [54], etc., have been widely applied to pattern recognition. The connecting weights are usually adjusted to minimize the squared error on training samples in supervised learning. Using a modular network for each class was shown to improve the classification accuracy [26]. A network using local connection and shared weights, called convolutional neural network, has reported great success in character recognition [27]. Kumar and Singh [12] proposed a Zernike moment feature based approach for Devnagari handwritten character recognition. They used an artificial neural network for classification. Bhattacharya et al [13,15] proposed a Multi-Layer Perceptron (MLP) neural network based classification approach for the recognition of Devnagari handwritten numerals. S.Arora [55] proposed a MLP designed on on some statistical features for handwritten devnagari characters recognition. The RBF network can yield competitive accuracy with the MLP when training all parameters by error minimization [28]. The HONN is also called as functional-link network, polynomial network or polynomial classifier (PC). Its complexity can be reduced by dimensionality reduction before polynomial expansion [29] or polynomial term selection [30]. Vector quantization (VQ) networks and auto-association networks, with the sub-net of each class trained independently in unsupervised learning, are also useful for classification. The learning vector quantization (LVQ) of Kohonen [31] is a supervised learning method and can give higher classification accuracy than VQ. Some improvements of LVQ learn prototypes by error minimization instead of heuristic adjustment [32].

### 3.3 Kernel based methods

Kernel based methods, including support vector machines (SVMs) [33, 34] primarily and kernel principal component analysis (KPCA), kernel Fisher discriminant analysis (KFDA), etc., are receiving increasing attention and have shown superior performance in pattern recognition. An SVM is a binary classifier with discriminant function being the weighted combination of kernel functions over all training samples. After learning by quadratic programming (QP), the samples of non-zero weights are called support vectors (SVs). For multi-class classification, binary SVMs are combined in either one-against-others or one-against-one (pair wise) scheme [35]. Due to the high complexity of training and execution, SVM classifiers have been mostly applied to small category set problems. A strategy to alleviate the computation cost is to use a statistical or neural classifier for selecting two candidate classes, which are then discriminated by SVM [30]. Dong et al. used a one-against-others scheme for large set Chinese character recognition with fast training [37]. They used a coarse classifier for acceleration but the large storage of SVs was not avoided.

### 3.4 Multiple classifier combination

Combining multiple classifiers has been long pursued for improving the accuracy of single classifiers [38]. Parallel (horizontal) combination is more often adopted for high accuracy, while sequential (cascaded, vertical) combination is mainly used for accelerating large category set classification. The decision fusion methods are categorized into abstract-level, rank-level, and measurement-level combination [39, 40]. Many fusion methods have been proposed [41, 42]. The complementariness (also called as independence or diversity) of classifiers is important to yield high combination performance. For character recognition, combining classifiers based on different techniques of pre-processing, feature extraction, and classifier models is effective. Bajaj et al [11] employed three features namely, density features, moment features and descriptive component features for classification of Devnagari Numerals. They proposed multi-classifier connectionist architecture for increasing the recognition reliability for handwritten Devnagari numerals. S. Arora et al [49] has worked on Chain code histogram, view based, intersection, shadow, momentum based and some curve fitting based features and classifiers combination method for handwritten Devnagari characters. Another effective method, called perturbation, uses a single classifier to classify multiple deformations of the input pattern and combine the decisions on multiple deformations [43, 44]. The deformations of training samples can also be used to train the classifier for higher generalization performance [44, 27].

## 4. Performance comparison

The experiments of character recognition reported in the literature vary in many factors such as the sample data, pre-processing technique, feature representation, classifier structure and learning algorithm. Only a few works have compared different classification/learning methods based on the same feature data. In the following, we first discuss the extracted feature data used for all classifiers, and then summarize some classification results on this feature data. As handwritten Devnagari characters have wide applicability in India, we tested the performance on it. As till date no standard dataset is available for Handwritten Devnagari Characters we collected some samples from ISI, Kolkata and some samples we created within our organization. Total of 7154 data samples are used out of which 4900 samples are provided by ISI, Kolkata. The database contains 4900 training samples and 2254 test samples. Each sample was normalized to binary image of 100 X 100 pixels and the features discussed below are extracted. These extracted features are used in MLP and SVM classifiers, discussed in section 5.

### 4.1 Shadow Features of character

Shadow is basically the length of the projection on the sides as shown in Figure 4. For computing shadow features [45] on scaled binary image, the rectangular boundary enclosing the character image is divided into eight octants. For each octant shadows or projections of character segment on three sides of the octant dividing triangles are computed so, a total of 24 shadow features are obtained. Each of these features is divided by the length of the corresponding side of the triangle to get a normalized value.

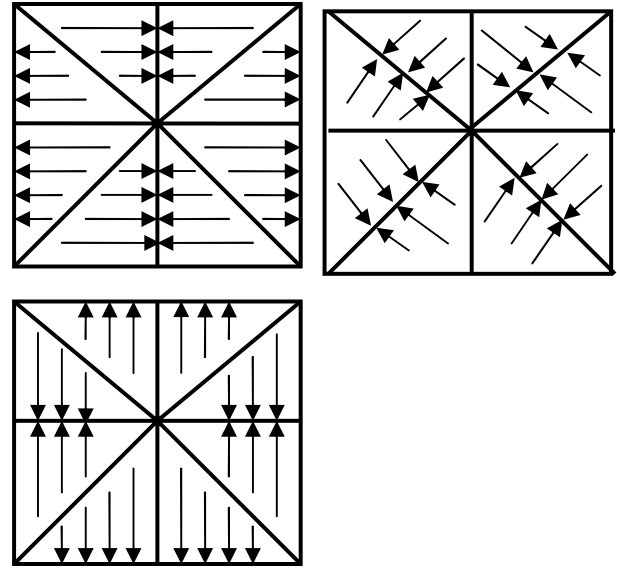

Figure 4. Shadow features

### 4.2 Chain Code Histogram of Character Contour

Given a scaled binary image, we first find the contour points of the character image. We consider a 3 × 3 window surrounded by the object points of the image. If any of the 4-connected neighbor points is a background point then the object point (P), as shown in Figure 5 is considered as contour point.

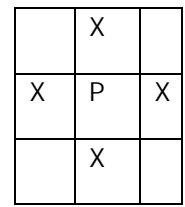

Figure 5. Contour point detection

The contour following procedure uses a contour representation called "chain coding" that is used for contour following proposed by Freeman[14], shown in Figure 6a. Each pixel of the contour is assigned a different code that indicates the direction of the next pixel that belongs to the contour in some given direction. Chain code provides the points in relative position to one another, independent of the coordinate system. In this methodology of using a chain coding of connecting neighboring contour pixels, the points and the outline coding are captured. Contour following procedure may proceed in clockwise or in counter clockwise direction.

Here, we have chosen to proceed in a clockwise direction.

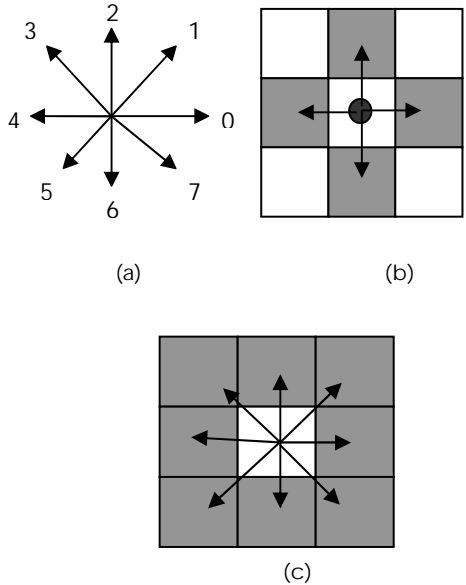

Figure 6. Chain Coding: (a) direction of connectivity, (b) 4-connectivity, (c) 8-connectivity. Generate the chain code by detecting the direction of the next-in-line pixel

The chain code for the character contour will yield a smooth, unbroken curve as it grows along the perimeter of the character and completely encompasses the character. When there is multiple connectivity in the character, then there can be multiple chain codes to represent the contour of the character. We chose to move with minimum chain code number first.
We divide the contour image in $5 \times 5$ blocks. In each of these blocks, the frequency of the direction code is computed and a histogram of chain code is prepared for each block. Thus for $5 \times 5$ blocks we get $5 \times 5 \times 8 = 200$ features for recognition.

### 4.3 View based features

This method is based on the fact, that for correct character-recognition a human usually needs only partial information about it – its shape and contour. This feature extraction method, which works on scaled, thinned binarized image, examines four "views" of each character extracting from them a characteristic vector, which describes the given character. The view is a set of points that plot one of four projections of the object (top, bottom, left and right) – it consists of pixels belonging to the contour of the character and having extreme values of one of its coordinates. For example, the top view of a letter is a set of points having maximal $y$ coordinate for a given $x$ coordinate. Next, characteristic points are marked out on the surface of each view to describe the shape of that view (Figure.7) The method of selecting these points and their number may vary and can be decided on experiment bases. In the considered examples, eleven uniformly distributed characteristic points are taken for each view.

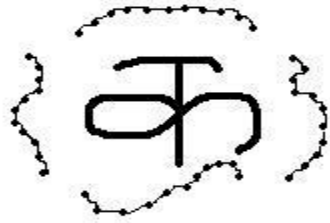

Figure 7. Selecting characteristic points for four views

The next step is calculating the $y$ coordinates for the points on the top and down views, and $x$ coordinates for the points on left and right views. These quantities are normalized so that their values are in the range <0, 1>. Now, from 44 obtained values the feature vector is created to describe the given character, and which is the base for further analysis and classification.

### 4.4 Longest-run Features

For computing longest-run features from a character image, the minimum square enclosing the image is divided into 25 rectangular regions. In each region, 4 longest-run features are computed row wise, column wise and along of its major diagonals. The row wise longest-run feature is computed by considering the sum of the lengths of the longest run bars that fit consecutive black pixels along each of all the rows of a rectangular region, as illustrated in Figure 8. The three other longest-run features are computed in the same way but along all column wise and two major diagonal wise directions within the rectangular separately. Thus in all, 25x4=100 longest-run features are computed from each character image.

Length of the Longest Bar

| 1 | 0 | 1 | 1 | 1 | 1 | 4 |
|---|---|---|---|---|---|---|
| 1 | 0 | 0 | 1 | 1 | 0 | 2 |
| 1 | 0 | 0 | 1 | 1 | 0 | 2 |
| 1 | 0 | 0 | 0 | 1 | 0 | 1 |
| 0 | 1 | 0 | 0 | 1 | 0 | 1 |
| 0 | 0 | 1 | 1 | 0 | 0 | 2 |

Sum=12

Figure 8: Longest run bar

## 5. Evaluated classifiers

### 5.1 Neural classifier

Combination of multiple classifiers is used to improve the accuracy in many pattern recognition tasks. We explored three ways of combining decision from multiple classifiers as a viable way of delivering a robust performance. All three ways are variations of majority voting scheme.

We designed the same MLP with 3 layers including one hidden layer for four different feature sets consisting of 100 longest run features, 24 shadow features, 44 view based features and 200 chain code histogram features. The classifier is trained with standard back propagation. It minimizes the sum of squared errors for the training samples by conducting a gradient descent search in the weight space. As activation function we used sigmoid function. Learning rate and momentum term are set to 0.8 and 0.7 respectively. As activation function we used the sigmoid function. Numbers of neurons in input layer of MLPs are 100, 24, 44 or 200, for longest run features, shadow features, view based features and chain code histogram features respectively. Number of neurons in Hidden layer is not fixed, we experimented on the values between 20-70 to get optimal result. The output layer contained one node for each class, so the number of neurons in output layer is 49. And classification was accomplished by a simple maximum response strategy.

### 5.1.1 Majority Voting

Majority Voting systems for decision combination, choices between selecting either the "consensus decision" or the "decision delivered by the most competent expert" strategy. This section presents some of the principal techniques based on Majority Voting System.

**Max Voting:** If there are $n$ independent experts having the same probability of being correct and each of these experts produce a unique decision regarding the identity of the unknown sample, then the sample is assigned to the class for which all $n$ experts agrees. Assuming that each expert makes a decision on an individual basis, without being influenced by any other expert in the decision making process.

**Min Voting:** If there are $n$ independent experts having the same probability of being correct and each of these experts produce a unique decision regarding the identity of the unknown sample, then the sample is assigned to the class for which any one of $n$ experts agrees. Assuming that each expert makes a decision on an individual basis, without being influenced by any other expert in the decision making process

### 5.1.2 Weighted Majority Voting

A simple enhancement to the simple majority systems can be made if the decisions of each classifier are multiplied by a weight to reflect the individual confidences of these decisions. In this case, Weighting Factors, $\omega_k$, expressing the comparative competence of the cooperating experts, are expressed as a list of fractions, with $1 \leq k \leq n$, $\sum_{k=1}^{n} \omega_k = 1$, $n$ being the number of participating experts. The higher the competence, the higher is the value of $\omega$. So if the decision by the $k^{th}$ expert to assign the unknown to the $i^{th}$ class is denoted by $d_{ik}$ with $1 \leq i \leq m$, $m$ being the number of classes, then the final combined decision $di^{com}$ supporting assignment to the $i^{th}$ class takes the form of: $di^{com} = \sum_{k=1,2,...n} \omega_k * d_{ik}$. The final decision $d^{com}$ is therefore: $d^{com} = \max_{i=1,2,...,m} d^{com}$. We worked on four features namely: Chain code histogram, shadow, view based and longest run features so we have **ω1, ω2, ω3** and **ω4** as 0.316, 0.303, 0.241 and 0.140 where $\omega_k$ is

$$\omega_k = \frac{d_k}{\sum_{k=1}^{4} d_k}$$

where **m** = 49 and $d_k$ is the result of classifiers trained with Chain code histogram, shadow based, view based features and longest run features.

### 5.2 Support Vector Machines

The objective of any machine capable of learning is to achieve good generalization performance, given a finite amount of training data, by striking a balance between the goodness of fit attained on a given training dataset and the ability of the machine to achieve error-free recognition on other datasets. With this concept as the basis, support vector machines have proved to achieve good generalization performance with no prior knowledge of the data. The principle of an SVM is to map the input data onto a higher dimensional feature space nonlinearly related to the input space and determine a separating hyperplane with maximum margin between the two classes in the feature space[47]. A support vector machine is a maximal margin hyperplane in feature space built by using a kernel function in gene space. This results in a nonlinear boundary in the input space. The optimal separating hyperplane can be determined without any computations in the higher dimensional feature space by using kernel functions in the input space. Commonly used kernels include:-

1. Linear Kernel :
$$K(x, y) = x.y$$
2. Radial Basis Function (Gaussian) Kernel :
$$K(x,y) = \exp(-||x - y||^2/2\sigma^2)$$
3. Polynomial Kernel :
$$K(x, y) = (x.y + 1)^d$$

An SVM in its elementary form can be used for binary classification. It may, however, be extended to multiclass problems using the one-against-the-rest approach or by using the one-against-one approach. We begin our experiment with SVM's that use the Linear Kernel because they are simple and can be computed quickly. There are no kernel parameter choices needed to create a linear SVM, but it is necessary to choose a value for the soft margin (C) in advance. Then, given training data with feature vectors $x_i$ assigned to class $y_i \in \{-1,1\}$ for $i = 1,……l$, the support vector machines solve

$$\min_{\xi, b, \omega} \frac{1}{2} K(\omega,\omega) + c \sum_{i=1}^{l} \xi_i$$

Subject to $y_i(K(\omega,x_i)+b) \geq 1 - \xi_i$
$\xi_i \geq 0$

where $\xi$ is an l-dimensional vector, and $\omega$ is a vector in the same feature space as the $x_i$. The values $\omega$ and $b$ determine a hyper plane in the original feature space, giving a linear classifier. A priori, one does not know which value of soft margin will yield the classifier with the best generalization ability. We optimize this choice for best performance on the selection portion of our data.

## 6. Performance Evaluation

We tested the performance on Handwritten Devnagari characters. As till date no standard dataset is available for Handwritten Devnagari Characters we collected some samples from ISI, Kolkata and some samples we created within our organization. Total of 7154 data samples are used out of which 4900 samples are provided by ISI, Kolkata. We tested Neural Networks and Support vector machines on ISI data samples and also on our own created data samples. For ISI dataset we considered 3430 data samples for training and 1470 for testing and for our own dataset of 2254 samples, we considered 1470 data samples for training and 784 for testing. Some samples of Devnagari characters are given in fig.9

Table 1. Results of SVM and ANN on ISI dataset

| Classifier | | Test set | Training set |
|---|---|---|---|
| **SVM** | | 80.67% | 94.77% |
| **Multiple Neural Network Classifier Combination (ANNs)** | Min | 60.92% | 76.75% |
| | Max | 74.65% | 84.12% |
| | Weighted majority | 70.38% (top1) 90.74% (top5) | 82.15%(top1) 93.31%(top5) |

Table 2. Results of SVM and ANN on Our dataset

| Classifier | | Test set | Training set |
|---|---|---|---|
| **SVM** | | 92.38% | 99.62% |
| **Multiple Neural Network Classifier Combination (ANNs)** | Min | 78.49% | 91.08% |
| | Max | 93.93% | 98.24% |
| | Weighted majority | 90.44% (top1) 99.08% (top5) | 97.94%(top1) 99.51%(top5) |

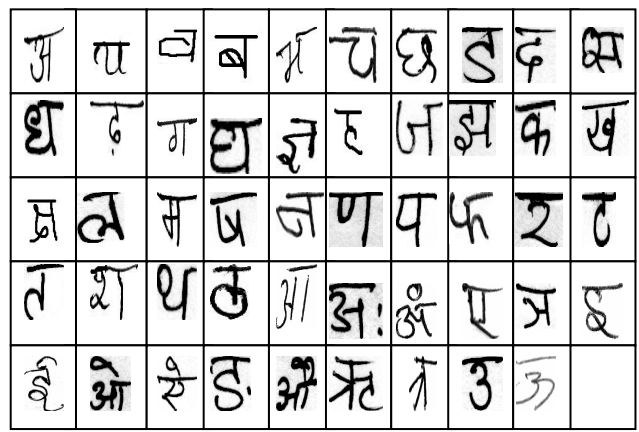

Figure 9: Some Devnagari Characters Sample set

## 7. Neural networks Vs. SVM

Neural classifiers and SVMs show different properties in the following respects.

*Complexity of training.* The parameters of neural classifiers are generally adjusted by gradient descent. By feeding the training samples a fixed number of sweeps, the training time is linear with the number of samples. SVMs are trained by quadratic programming (QP), and the training time is generally proportional to the square of number of samples. Some fast SVM training algorithms with nearly linear complexity are available, however.

*Flexibility of training.* The parameters of neural classifiers can be adjusted in string-level or layout-level training by gradient descent with the aim of optimizing the global performance [48]. In this case, the neural classifier is embedded in the string or layout recognizer for character recognition. On the other hand, SVMs can only be trained at the level of holistic patterns.

*Model selection.* The generalization performance of neural classifiers is sensitive to the size of structure, and the selection of an appropriate structure relies on cross-validation. The convergence of neural network training suffers from local minima of error surface. On the other hand, the QP learning of SVMs guarantees finding the global optimum. The performance of SVMs depends on the selection of kernel type and kernel parameters, but this dependence is less influential.

*Classification accuracy.* SVMs have been demonstrated superior classification accuracies to neural classifiers in many experiments.

*Storage and execution complexity.* SVM learning by QP often results in a large number of SVs, which should be stored and computed in classification. Neural classifiers have much less parameters, and the number of parameters is easy to control. In a word, neural classifiers consume less storage and computation than SVMs.

## 8. Conclusion

The result obtained for recognition of Devnagari characters show that reliable classification is possible using SVMs. We applied SVMs and ANNs classifiers on same feature data namely Shadow based, Chain code Histogram, Longest Run, and View based features. The SVM-based method described here for offline Devnagari can be easily extended to other Indian scripts and Handwritten Devnagari numerals also.

**Acknowledgments**

Authors are thankful to the "Centre for Microprocessor Application for Training Education and Research" and "Project on Storage Retrieval and Understanding of Video for Multimedia", at the Department of Computer Science and Engineering, Jadavpur University, Kolkata-700032 for providing the necessary facilities for carrying out this work. Authors are also thankful to the CVPR Unit, ISI Kolkata for providing the dataset of Handwritten Devnagari Characters. First author gratefully acknowledge the support of the Meghnad Saha Institute of Technology for carrying out this research work.

**SANDHYA ARORA** has completed M.Tech. (Computer Science & Engineering) from Banasthali Vidyapith, Rajasthan, India and B.E.(Computer Engineering) from University of Rajasthan, India .She is currently working as Assistant Professor in Department of Computer Science & Engineering at Meghnad Saha Institute of Technology, Kolkata, WB, India. She has teaching experience of 13 years. Mrs.Sandhya Arora presented 5 papers in national and 4 papers in international conferences She is life member of CSI.

**DEBOTOSH BHATTACHARJEE** received the MCSE and Ph.D. (Eng.) degrees from Jadavpur University, India, in 1997 and 2004 respectively. He was associated with different institutes in various capacities until March 2007. After that he joined his Alma Mater, Jadavpur University. His research interests pertain to the applications of computational intelligence techniques like Fuzzy logic, Artificial Neural Network, Genetic Algorithm, Rough Set Theory, Cellular Automata etc. in Face Recognition, OCR, and Information Security. He is a life member of Indian Society for Technical Education (ISTE, New Delhi), Indian Unit for Pattern Recognition and Artificial Intelligence (IUPRAI), and member of IEEE (USA).

**MITA NASIPURI** received his B.E.Tel.E., M.E.Tel.E. and Ph.D. (Eng.) degrees from Jadavpur University, in 1979,1981 and 1990 respectively. Prof. Masipuri has been a faculty member of J.U. since 1987. Her current research interest include pattern recognition, image processing and multimedia systems. She is a senior member of the IEEE, U.S.A., Fellow of I.E. (India) and W.B.A.S.T., kolkata , India.

**MAHANTAPAS KUNDU** received his B.E.E., M.E.Tel.E. and Ph.D.(Eng.) degrees from Jadavpur University, in 1983,1985 and 1995respectively. Prof. Kundu has been a faculty member of J.U. since 1988. His areas of current research interest include pattern recognition, image processing , multimedia database, and artificial intelligence.

**DIPAK KUMAR BASU** received his B.E.Tel.E., M.E.Tel., and Ph.D. (Eng.) degrees from Jadavpur University, in 1964,1966 and 1969 respectively. Prof. Basu has been a faculty member of J.U.since 1968. His current fields of research interest include pattern recognition, image processing, and multimedia systems. He is a senior member of the IEEE, U.S.A., Fellow of I.E. (India) and W.B.A.S.T., kolkata , India and a former Fellow, Alexander von Humboldt Foundation, Germany.

**LATESH MALIK** became a Member (M) of IEEE in 2006. She has completed M.Tech. (Computer Science & Engineering) from Banasthali Vidyapith, Rajasthan, India and B.E. (Computer Engineering) from University of Rajasthan,India . She is gold medallist in B.E. and M.Tech. She is currently working as Assistant Professor & Head of Department in Department of Computer Science & Engineering at G.H. Raisoni College of Engineering, Nagpur University, Nagpur, MS, India. She has teaching experience of 13 years.Mrs. Latesh Malik is life member of ISTE and presented 5 papers in international journal, 8 papers in national and 20 papers in international conferences.